\documentclass[acmtog]{acmart}
\AtBeginDocument{%
  }

\setcopyright{acmlicensed}
\copyrightyear{2024}
\acmYear{2024}

\usepackage[capitalise]{cleveref}
\usepackage{graphicx}
\usepackage{subfig}
\usepackage{booktabs}
\usepackage{lscape}
\usepackage{longtable}
\usepackage{caption}
\usepackage{float}
\usepackage{array}
\usepackage{multirow}
\begin{document}

\title{Introducing a new hyper-parameter for RAG: Context Window Utilization}

\author{Kush Juvekar}
\email{kushjuveka0610@gmail.com}
\affiliation{%
  \institution{Independent}
  \city{Ahmedabad}
  \state{Gujarat}
  \country{India}
}
\author{Anupam Purwar}
\email{anupam.aiml@gmail.com}
\orcid{0000-0002-4484-7575}
\affiliation{%
  \institution{Independent}
  \city{Delhi}
  \country{India}
}

\begin{abstract}
This paper introduces a new hyper-parameter for Retrieval-Augmented Generation (RAG) systems called Context Window Utilization. RAG systems enhance generative models by incorporating relevant information retrieved from external knowledge bases, improving the factual accuracy and contextual relevance of generated responses. The size of the text chunks retrieved and processed is a critical factor influencing RAG performance. This study aims to identify the optimal chunk size that maximizes answer generation quality. Through systematic experimentation, we analyze the effects of varying chunk sizes on the efficiency and effectiveness of RAG frameworks. Our findings reveal that an optimal chunk size balances the trade-off between providing sufficient context and minimizing irrelevant information. These insights are crucial for enhancing the design and implementation of RAG systems, underscoring the importance of selecting an appropriate chunk size to achieve superior performance.
\end{abstract}



\maketitle

\section{Introduction}
Retrieval-Augmented Generation (RAG)\cite{lewis2021retrievalaugmentedgenerationknowledgeintensivenlp} is the process of optimizing the output of a large language model, so it references an authoritative knowledge base outside of its training data sources before generating a response. Large Language Models (LLMs) are trained on vast volumes of data and use billions of parameters to generate original output for tasks like answering questions, translating languages, and completing sentences. RAG extends the already powerful capabilities of LLMs to specific domains or an organization's internal knowledge base, all without the need to retrain the model. Implementing RAG is an efficient way to enhance the output of LLM, ensuring its relevance, accuracy, and usefulness in diverse applications like as \href{http://52.0.160.0}{chat-bots} for enterprise data \cite{juvekar2024cosmixcosinesimilaritydistance}, conversational learning interfaces \cite{harbola2024vidyarangconversationallearningbased}, and seamless integration with voice interfaces \cite{purwar2023keywordaugmentedretrievalnovel}.

Currently, most RAG systems follow the Dual-Encoder Architecture paradigm\cite{dong2022exploringdualencoderarchitectures}, in which the reference documents are divided into small snippets, encoded by specific encoders, and then stored in the vector database as embeddings. The DEA paradigm shows great potential for empowering LLMs with any existing knowledge database.

Present day advancements in RAG have answered various questions like "what should be retrieved" , "when retrieval should not occur", "how to utilize the retrieved information" and more.\cite{asai2023selfraglearningretrievegenerate,yan2024correctiveretrievalaugmentedgeneration,githubGitHubMicrosoftgraphrag,ganesh2024contextaugmentedretrievalnovelframework}.

However, the  Failure point 1(Missing Context) as articulated by
\cite{barnett2024sevenfailurepointsengineering} is often a result of inefficient chunking and indexing strategies. Many researchers have tried to address this by introducing advanced chunking strategies, both heuristic and semantic \cite{chen2023densexretrievalretrieval,duarte2024lumberchunker,bitpeakBitPeak,ChunkingStrategies}. Despite all the research in this area, determining an optimal \textbf{chunk size} remains a tedious task as user queries differ in nature leading to answers of varied nature dependant on tasks like summarization, fact finding, translation and question answering tasks. 

Our current work address the issue of finding an optimal chunk size for naive RAG (fixed chunk size couple with a dense retriever) in question answering task. The motivation is to find a chunk size optimal for all tasks then the other RAG advancements will certainly help RAG perform better on top of this. We demonstrate that, while documents may have distinct text alignment structures, LLMs do not respond in dramatically different ways. Through extensive experimentation with different chunk sizes across diverse datasets, including academic papers, legal documents, and Wikipedia articles, we aim to identify an optimal chunk size that balances the trade-off between providing sufficient context and minimizing irrelevant information. Our study introduces a new parameter, Context Window Utilization, which optimizes the proportion of the context window used during retrieval. Our findings based on experimenting with 2 different open source LLMs offer valuable insights into how to improve the design and implementation of RAG systems by using most optimal chunk size.

\section{Methodology}
\label{methodology}
\subsection{Dataset Preparation}
To explore the practical applications of Retrieval-Augmented Generation (RAG) with Large Language Models (LLMs), we conducted our testing using a diverse set of data corpora. These included scholarly publications, news articles, reports from the Indian government, newsletters, and decisions from the Indian High Court and Supreme Court. These document types are among the most frequently used documents by individuals in their daily activities. 

\begin{table*}
\centering
\begin{tabular*}{\textwidth}{@{\extracolsep{\fill}}|l|p{6.5cm}|p{6.5cm}|@{}}
\toprule
Dataset & Question & Answer \\ \midrule
\multirow{2}{*}{MS MARCO} & the meaning of reverence & Deep respect \\ \cmidrule(l){2-3} 
& define traumatic & It is extremely upsetting or something that causes great harm or damage to the mind or body. \\ \midrule
\multirow{2}{*}{NarrativeQA} & What is the name of the night club in Los Angeles where the Doors grew their fan base? & Whiskey a Go Go \\ \cmidrule(l){2-3} 
& What does Corso realize when he compares Balkan's and Fargas' copies of the Ninth Gate? & Each book has three different unique engravings with Lucifer's signature. \\ \midrule
\multirow{2}{*}{Natural Questions} & which type of rock forms on the earth's crust & Igneous and metamorphic rocks make up 90-95\% of the 16km of the Earth's crust by volume. Igneous rocks form about 15\% of the Earth's crust by volume. Most of the Earth's oceanic crust is made of igneous rock. \\ \cmidrule(l){2-3} 
& who was the first person to see earth from space & \begin{tabular}[c]{@{}p{6.5cm}@{}}Yuri Alekseyevich Gagarin was a Soviet pilot and cosmonaut. He was the first human to journey into outer space when his Vostok spacecraft completed an orbit of the Earth on 12 April 1961.\end{tabular}  \\ \bottomrule
\end{tabular*}
\caption{This table presents a random collection of questions and answers extracted from three datasets: MS MARCO, NarrativeQA, Natural Questions.}
\label{tab:qa_datasets}
\end{table*}

For each corpus, vector databases were constructed with varying chunk sizes of 128, 256, 512, 1024, and 2048. We employed a sentence splitter for segmentation and embedded the text using the Sentence Transformer model\cite{bge_embedding}.

We have utilized GPT-4 Turbo \cite{gpt4turbo} to generate question-answer pairs for each corpus. GPT-4 Turbo was selected based on evidence indicating that large language models achieve an 80\% alignment with human judgment\cite{zheng2023judgingllmasajudgemtbenchchatbot}. Synthetic data generation using LLMs has proven to be useful in many cases for example, 83\% GPT-4 generated example-label pairs were found to be correct by a clinician in this study\cite{woolsey2024utilizinglargelanguagemodels}. As LLMs are a probabilistic model working the principle of next token prediction they are prone to hallucinations. But, their ability to align with human preferences over 80\% of the time demonstrates their potential in tasks requiring natural language understanding and generation.

\subsection{Why custom datasets?}

We opted not to use existing datasets such as Natural Questions \cite{kwiatkowski-etal-2019-natural}, MS MARCO\cite{bajaj2018msmarcohumangenerated}, and BEIR \cite{thakur2021beirheterogenousbenchmarkzeroshot} due to several limitations. These datasets typically contain straightforward and direct queries, resulting in high semantic overlap between the queries and documents. In contrast, our queries encompass a mix of "What?", "How?", and "Why?" questions, which are more complex and diverse than simple factual questions. We have instructed GPT-4 Turbo to generate such queries as such queries are of greater relevance to a user rather than fact finding.

\cref{tab:qa_datasets} displays  randomly selected set of question answer pairs from frequently used datasets in RAG assessment. As evident from \cref{tab:qa_datasets}, all inquiries are quite direct and do not need the LLMs to engage in reasoning, but rather to simply search for factual information. Due to the factual nature of queries in datasets like MS MARCO, we created our own dataset which has a lot of descriptive inquiries. Some of the question answer pairs from our datasets are described in \cref{A}.

Another justification for avoiding the usage of these datasets is because text embedding models have been fine-tuned using them.  The sentence transformer utilized in this research is trained on many datasets including MS MARCO, NQ, HotpotQA, and others \cite{chen2024bgem3embeddingmultilingualmultifunctionality}. Additionally, another model \textit{gte-large-en-v1.5} has been trained on the MS MARCO, NQ, and FEVER among other datasets. Both models are on the top of the MTEB leaderboard among models that have fewer than 500 million parameters \cite{muennighoff2023mtebmassivetextembedding}. Therefore, evaluating the RAG on these datasets would result in good results as the embedding models have already learned the vector representations of these datasets.

\subsection{Evaluation}
To assess the quality of the generated response, we used open-source models viz. Llama 3 and Mixtral-8X7b to generate responses for the same set of questions. We then assessed the semantic similarity between GPT-4 Turbo's ground truth responses and the answers generated by open-source models. The semantic similarity score has been used to evaluate the quality of the responses, as this metric has demonstrated its efficacy in measuring the quality of response from LLMs.\cite{b2024evaluatingefficacyopensourcellms,risch2021semanticanswersimilarityevaluating}

\section{Results and Discussion}
\label{resultsanddiscussion}

We use two open source large language models(LLMs): Llama3-70B-Instuct\cite{llama3modelcard} and Mixtral-8x7B-Instruct-v0.1\cite{jiang2024mixtralexperts} for evaluating the impact of chunk size on quality of response. We calculated response quality for three types of data corpora: Wikipedia articles, legal documents and research papers to comprehensively cover all different document types and their associated structure.

\begin{landscape}
\begin{figure*}[!htb]
    \centering
    \begin{minipage}{0.49\linewidth}
        \centering
        \subfloat[Legal Documents (llama3)]{%
            \includegraphics[width=0.8\linewidth,height=0.3\textheight,keepaspectratio]{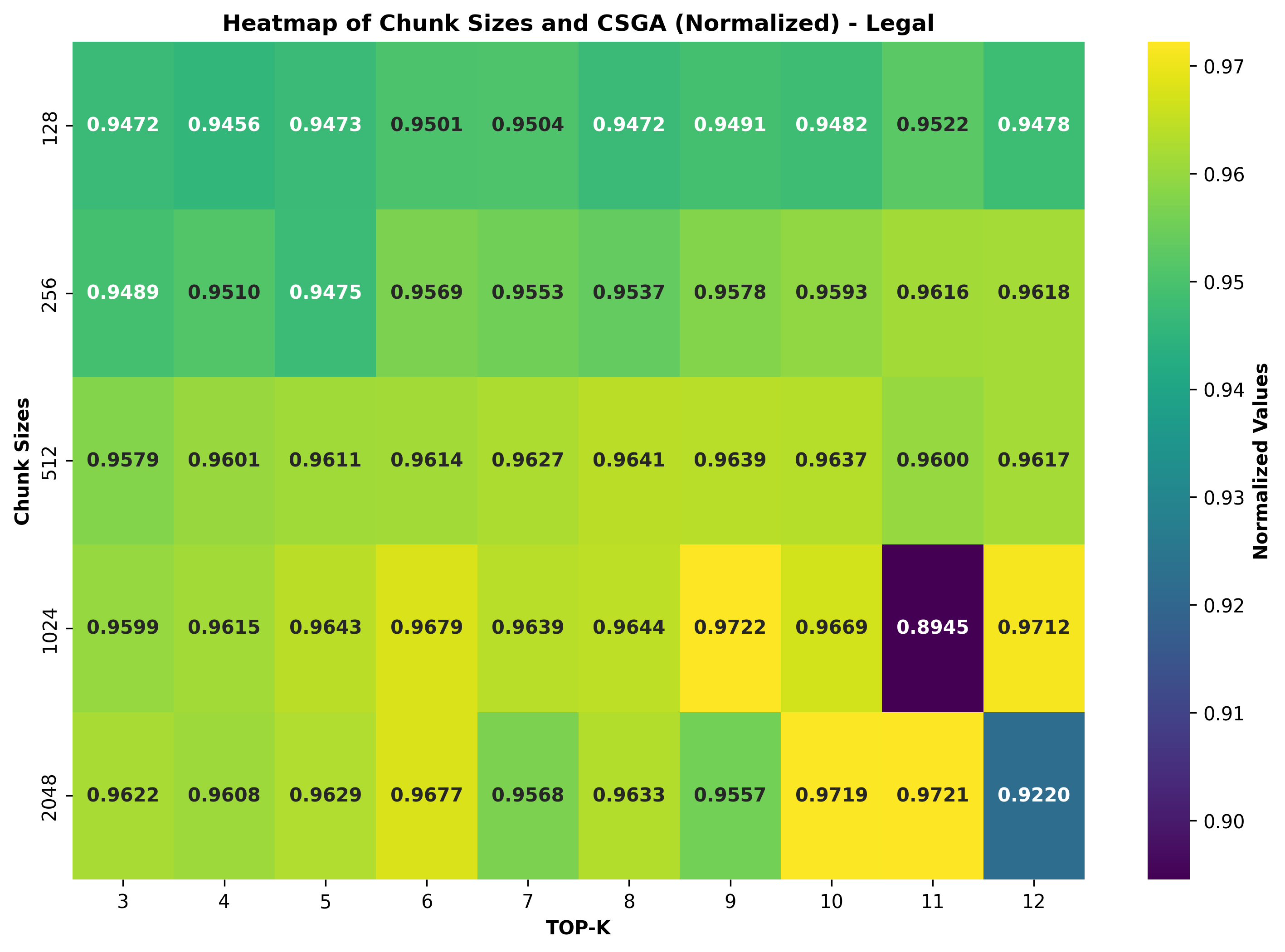}%
            \label{fig:image1a}%
        }
    \end{minipage}%
    \hfill
    \begin{minipage}{0.49\linewidth}
        \centering
        \subfloat[Legal Documents (mixtral)]{%
            \includegraphics[width=0.8\linewidth,height=0.3\textheight,keepaspectratio]{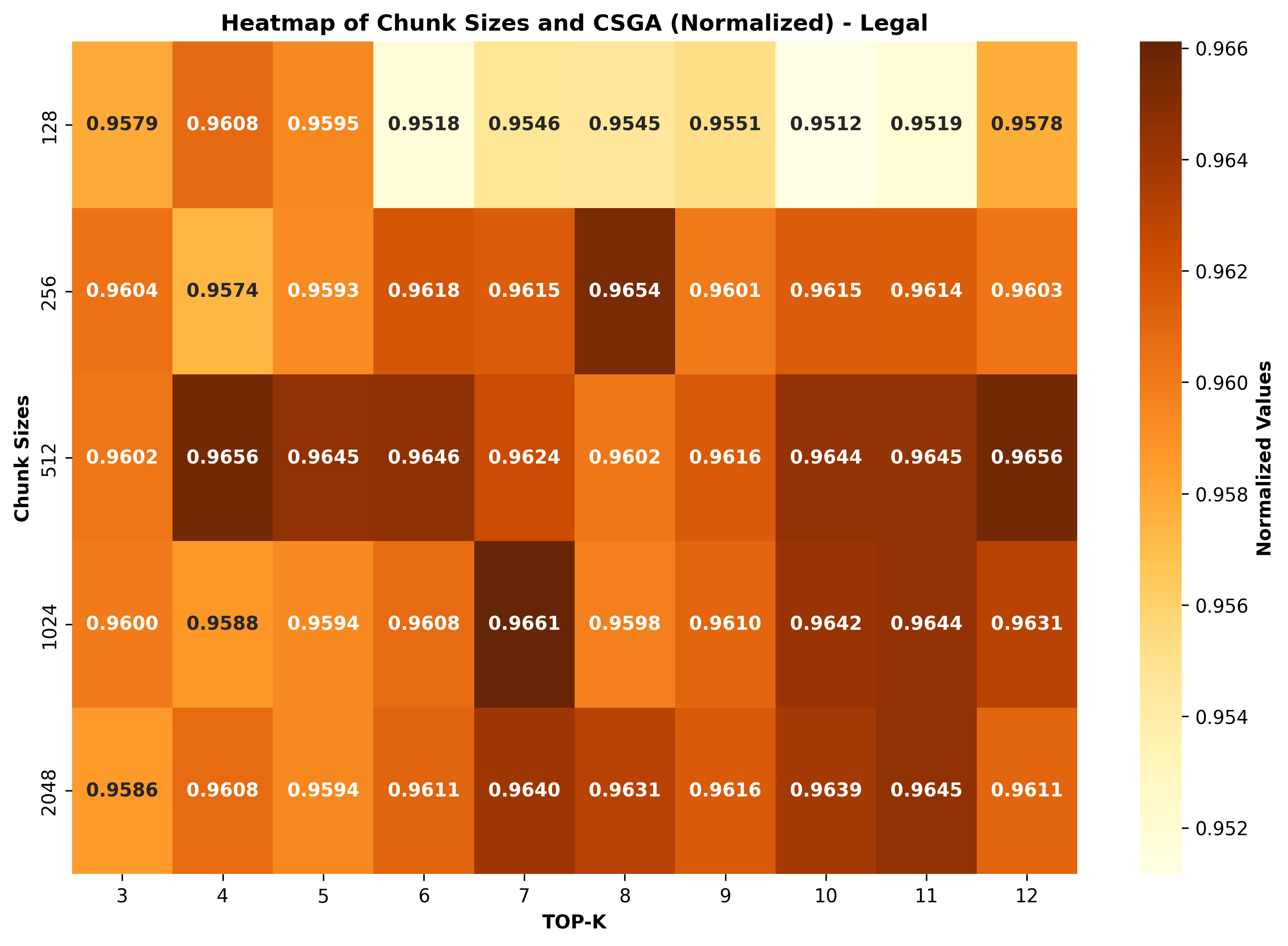}%
            \label{fig:image2b}%
        }
    \end{minipage}
    \vfill
    \begin{minipage}{0.49\linewidth}
        \centering
        \subfloat[Research Papers (llama3)]{%
            \includegraphics[width=0.8\linewidth,height=0.3\textheight,keepaspectratio]{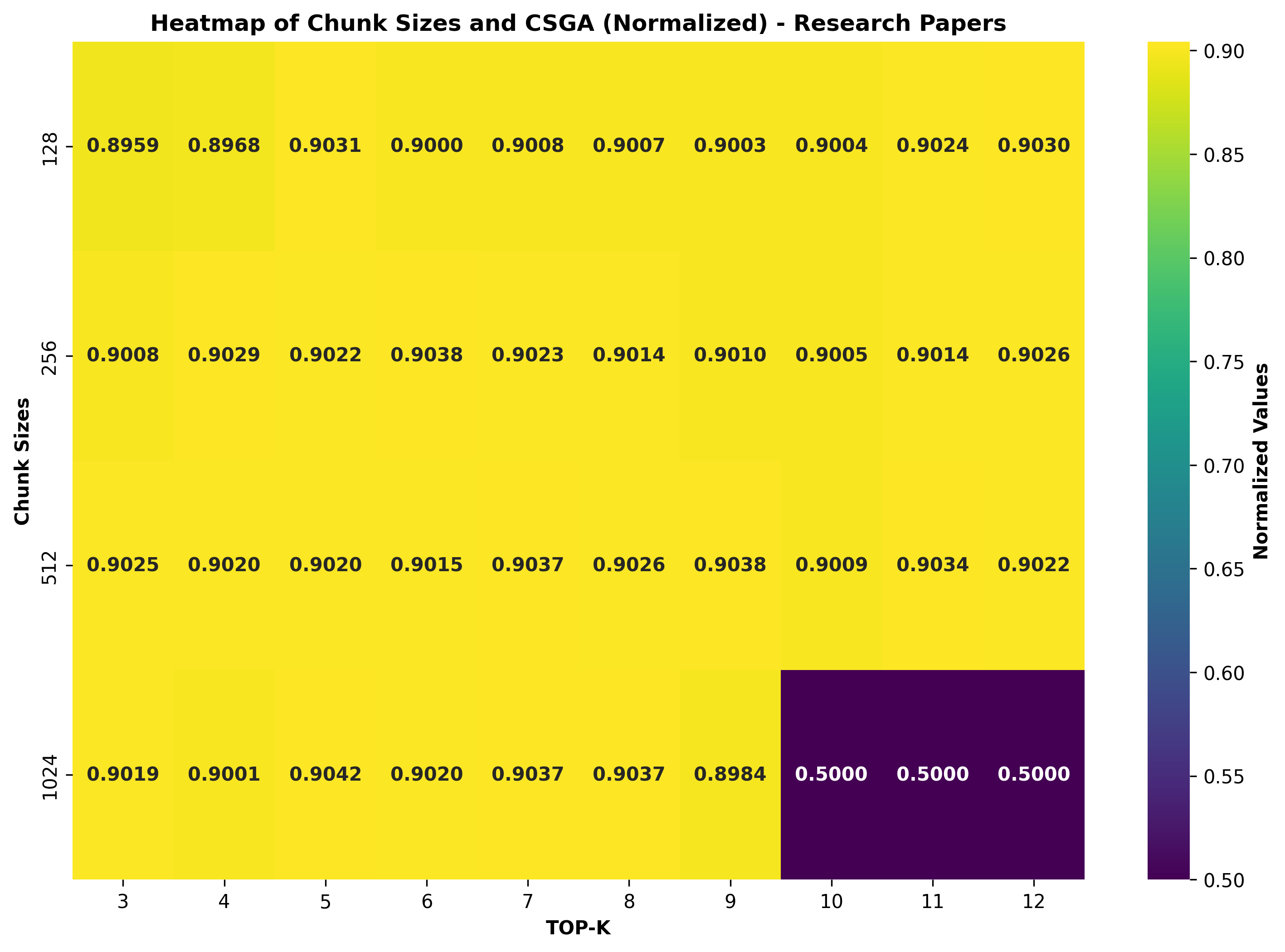}%
            \label{fig:image2a}%
        }
    \end{minipage}%
    \hfill
    \begin{minipage}{0.49\linewidth}
        \centering
        \subfloat[Research Papers (mixtral)]{%
            \includegraphics[width=0.8\linewidth,height=0.3\textheight,keepaspectratio]{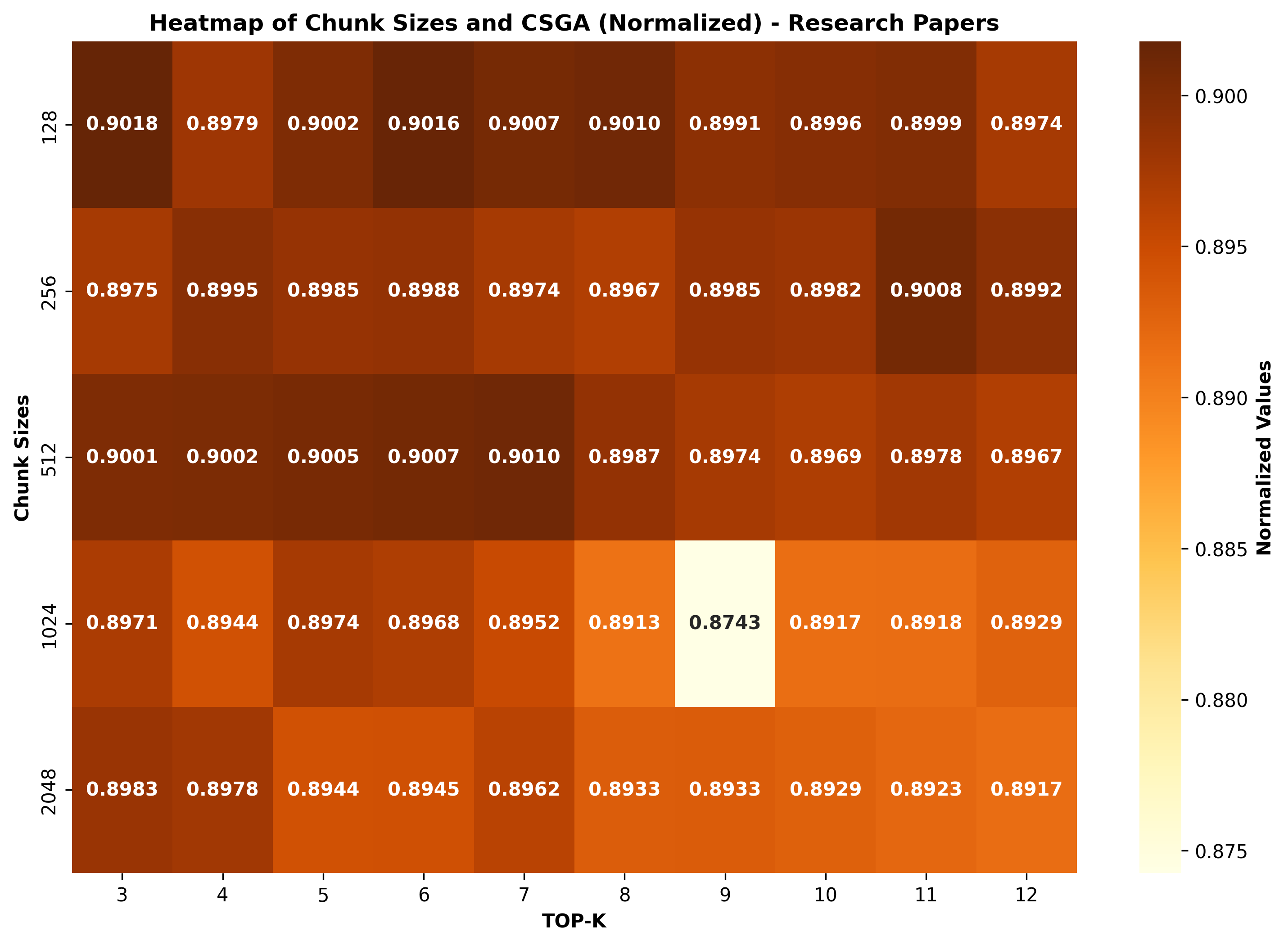}%
            \label{fig:image1c}%
        }
    \end{minipage}
    \vfill
    \begin{minipage}{0.49\linewidth}
        \centering
        \subfloat[Wikipedia Pages (llama3)]{%
            \includegraphics[width=0.8\linewidth,height=0.3\textheight,keepaspectratio]{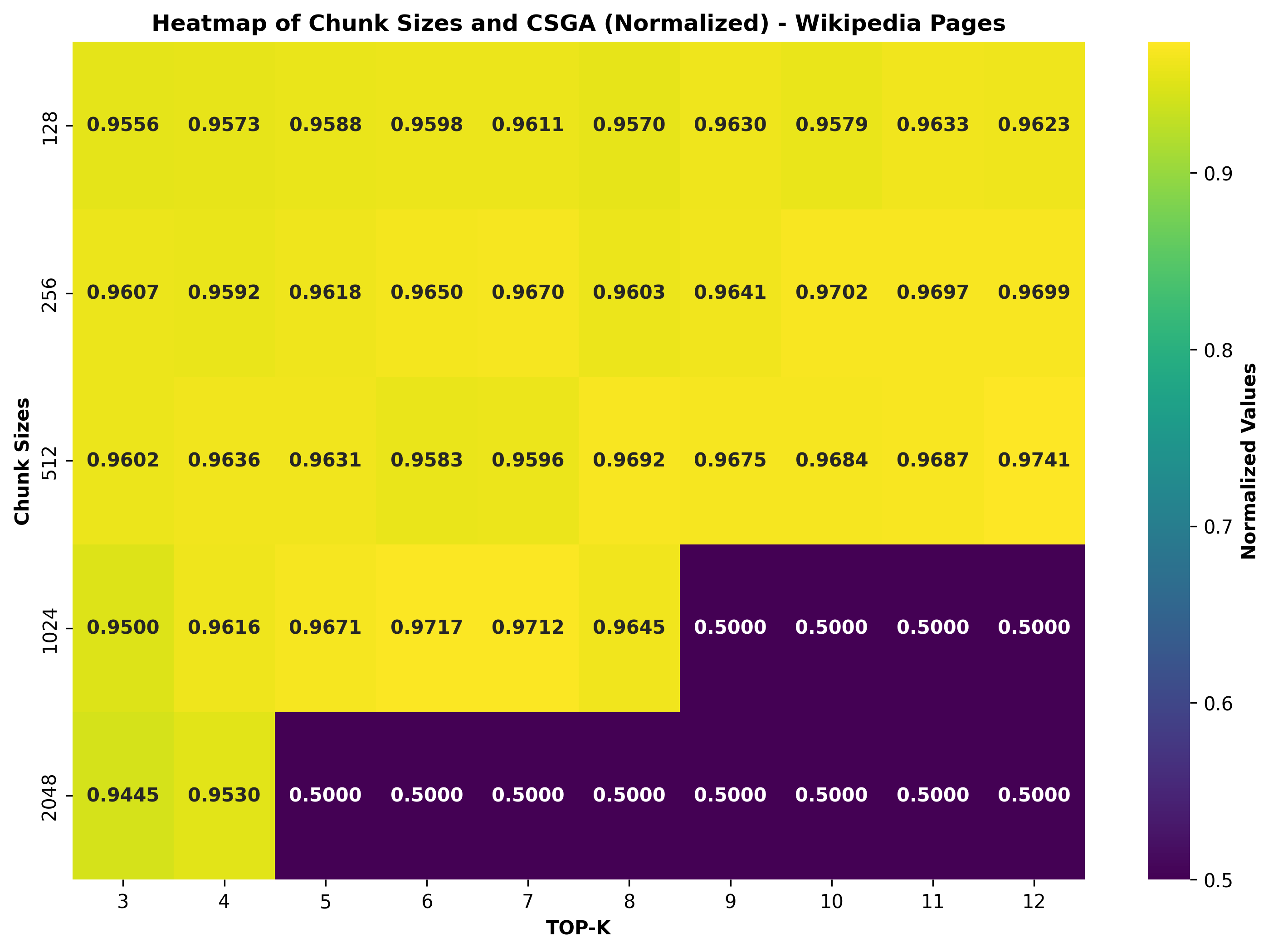}%
            \label{fig:image1b}%
        }
    \end{minipage}%
    \hfill
    \begin{minipage}{0.49\linewidth}
        \centering
        \subfloat[Wikipedia Pages (mixtral)]{%
            \includegraphics[width=1.5\linewidth,height=0.3\textheight,keepaspectratio]{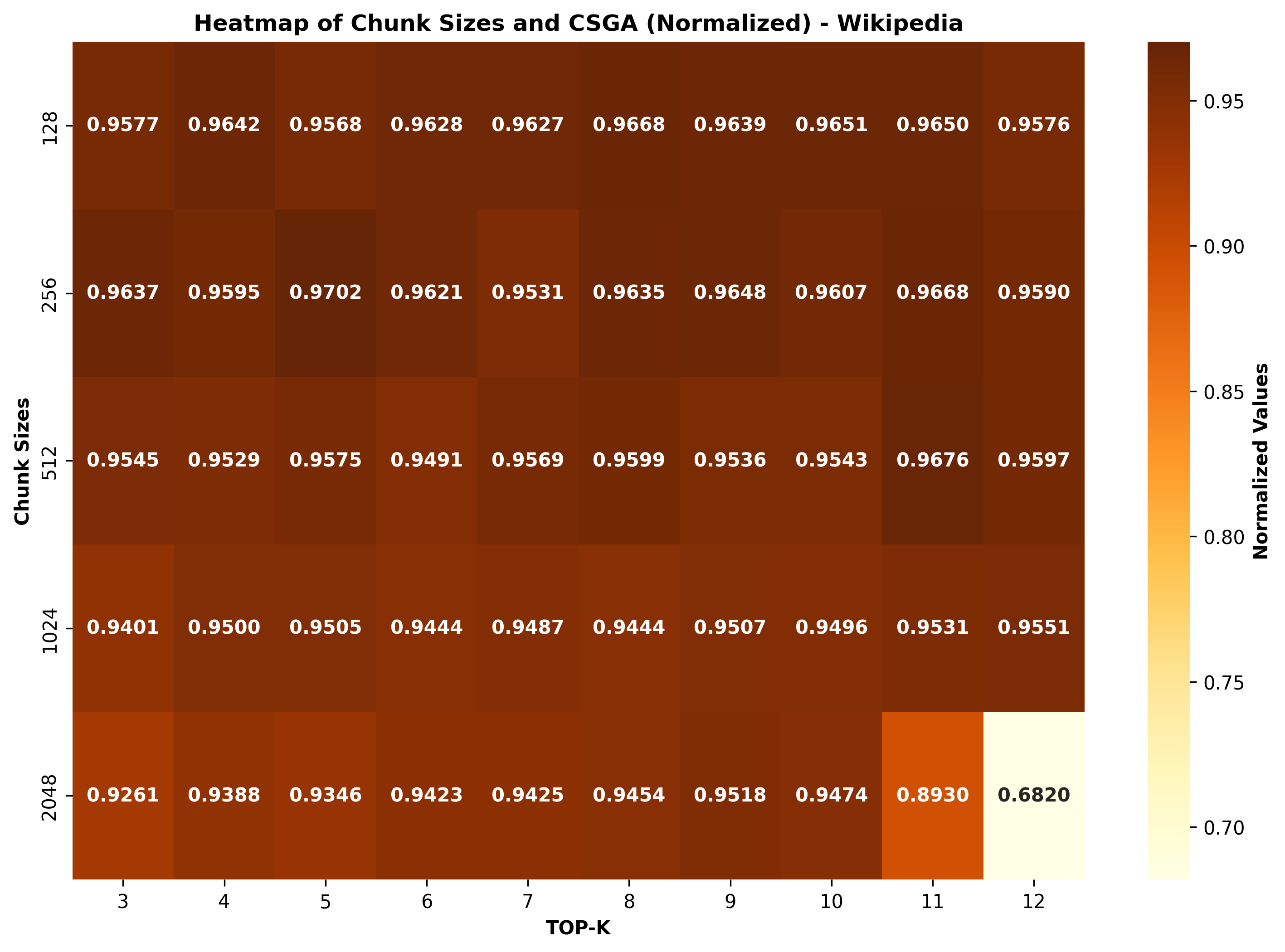}%
            \label{fig:image2c}%
        }
    \end{minipage}
    \caption{Semantic Similarity Comparisons}
    \label{Fig: Fig.1}
\end{figure*}
\end{landscape}
The table below shows the pair of chunk size and top-k where the similarity scores are maximized for each dataset for both models.
\begin{table}[h!]
\centering
\setlength{\extrarowheight}{-5pt} 
\resizebox{\columnwidth}{!}{%
\begin{tabular}{@{}lccc@{}}
\toprule
\textbf{Dataset} & \multicolumn{1}{l}{\textbf{Chunk Size}} & \multicolumn{1}{l}{\textbf{Top-K}} & \multicolumn{1}{l}{\textbf{Similarity Score}} \\ \midrule
\multicolumn{4}{c}{\textbf{Llama3-70B-Instruct}}                                                                                                \\
Wikipedia Pages  & 512                                     & 12                                 & 0.9741                                        \\
Legal Documents  & 1024                                   & 9                                  & 0.9722                                        \\
Research Papers  & 1024                                    & 5                                  & 0.9042                                        \\
\multicolumn{4}{c}{\textbf{Mixtral-8x7B-Instruct-v0.1}}                                                                                         \\
Wikipedia Pages  & 512                                     & 12                                 & 0.9597                                        \\
Legal Documents  & 1024                                    & 7                                  & 0.9661                                        \\
Research Papers  & 128/512\footnotemark[1]                                    & 3/7                                  & 0.9018/0.9010                                       \\ \bottomrule
\end{tabular}%
}
\end{table}
\footnotetext[1]{Both chunk sizes 128 and 512 give comparable performance here.
}

While a spike in similarity scores can be found at arbitrary combination of chunk size and top-k, we find that across all three document types chunk sizes 512 or 1024 consistently outperform other chunk sizes for both models (\cref{Fig: Fig.1}).

With Mixtral-8x7B-Instruct-v0.1, we find inconsistency in scores. This may arise due to the fact that its context window is 32K tokens,
but we have limited ourselves to 12 chunks at the maximum, thus utilizing only a small fraction of the large input space it has to offer. Other reason could be its unique architecture i.e Mixture of Experts which responds differently as compared to base or instruct models in retrieval augmented generation tasks . A recent blog by NVIDIA shows 8x7B doesn't mean 56B parameters but 47B due to shared layers and despite load balancing algorithms there are large distribution imbalances in tokens assigned to each expert causing some experts to finish early while others are overloaded \cite{nvidiaApplyingMixture}.

The Llama3-70B-Instruct model has the maximum similarity scores when there are 7 to 9 chunks with chunk sizes of 512 and 1024. This corresponds to a context window occupancy of 40-70\%. Furthermore, there is no improvement in the similarity score generated beyond 10 chunks in almost all combinations. This begs the question that along with an optimal chunk size, \textbf{utilization} of context window might also be an important factor. Similarity score is maximized at 40-50\% of the context window for 512 chunk size and 60-70\% for 1024 chunk size.

The considerably better results on the Wikipedia dataset for both LLMs are most likely attributable to the fact that Wikipedia pages are a typical element of the LLM pre-training process. While this maybe viewed as a limitation, there is no way to certainly ensure what data is used in pre-training LLMs. Researchers claim that publicly available resources and internet-scale data are used to train these LLMs \cite{metaIntroducingMeta,jiang2024mixtralexperts}. Popular datasets like  MS MARCO, Natural Questions, BEIR etc. could also well be part of their pre-training process.

\begin{figure}[h!]
  \centering
  \includegraphics[width=\columnwidth]{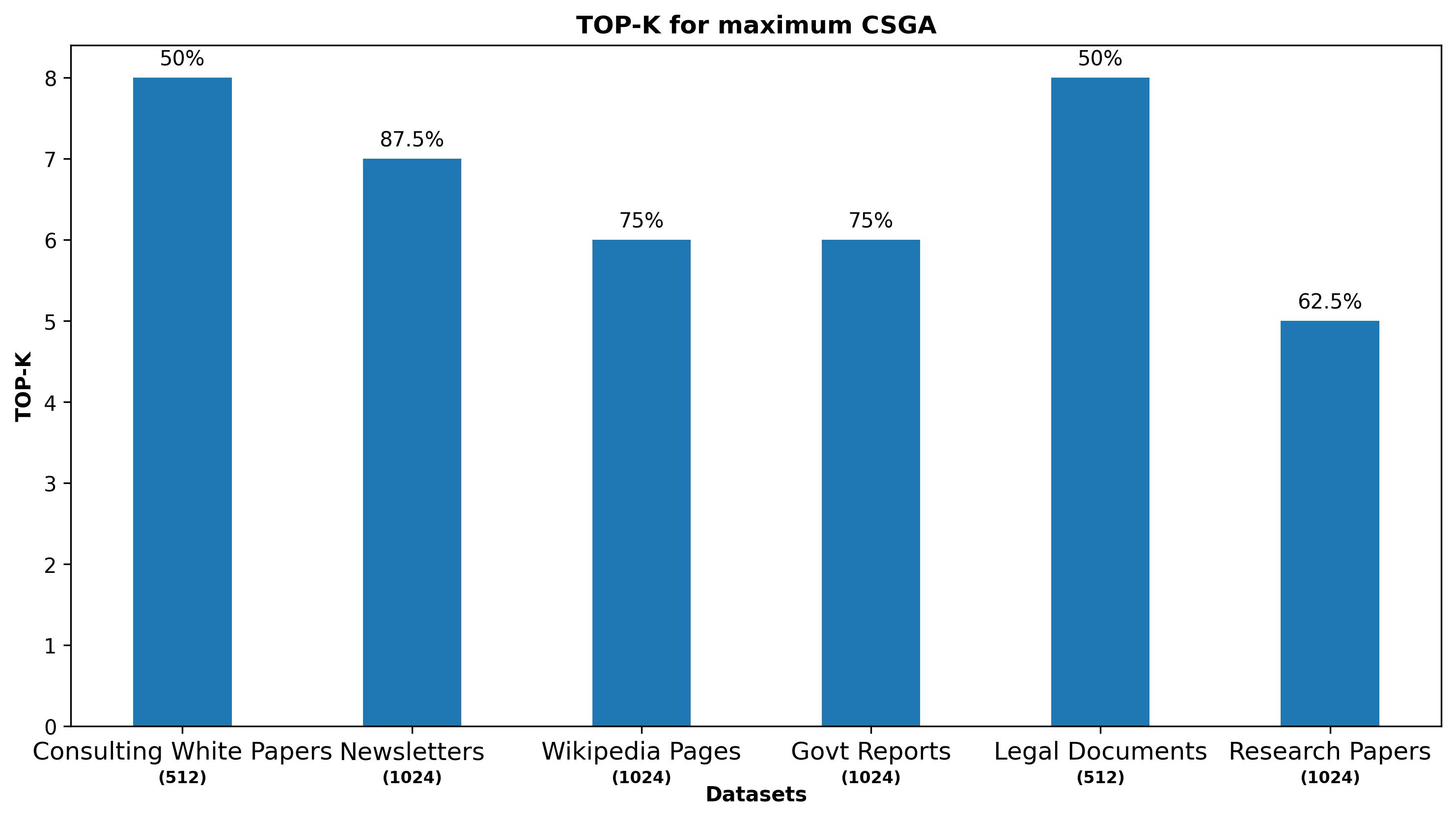}
  \caption{Number of chunks for which semantic similarity with GPT-4 Turbo answers is maximum in the distinct datasets. The chunk size for each dataset where scores are maximized are also mentioned in brackets.}
  \label{fig: Fig.2}
\end{figure}

Furthermore, we average the scores for all questions to determine which value of top-k has the highest value (\cref{fig: Fig.2}). With the exception of academic publications, all datasets scores are maximum for chunk numbers between 6 and 9. The percentage of context window utilized for maximum scores are also displayed at the top of each bar in the same figure. This supports our hypothesis  that there is an ideal context window utilization percentage where the quality of response is best. Thus, LLM context window utilization is an important parameter to be kept in mind while optimizing the performance of RAG workflow.

\textbf{Note:} A value of 0.5 on the heatmaps (\cref{Fig: Fig.1}) either means chunks retrieved are over the limit of LLM's context window or the API through which LLM is inferenced threw an unexpected error \href{https://developer.mozilla.org/en-US/docs/Web/HTTP/Status/400}{(400 Bad Request)}. 

\subsection{Context Window Utilization}

Our experiment with Llama3-70b-Instruct model demonstrates that Context window is an important parameter to be kept in mind while building the RAG pipeline. It can be defined as follows: \[
\text{CWU} = \frac{U}{L}
\] where U is the number of tokens actually utilized by the model within the context window and
L is the length of the context window (in tokens). 

\begin{figure*}[h!]
  \centering
  \includegraphics[width=\textwidth]{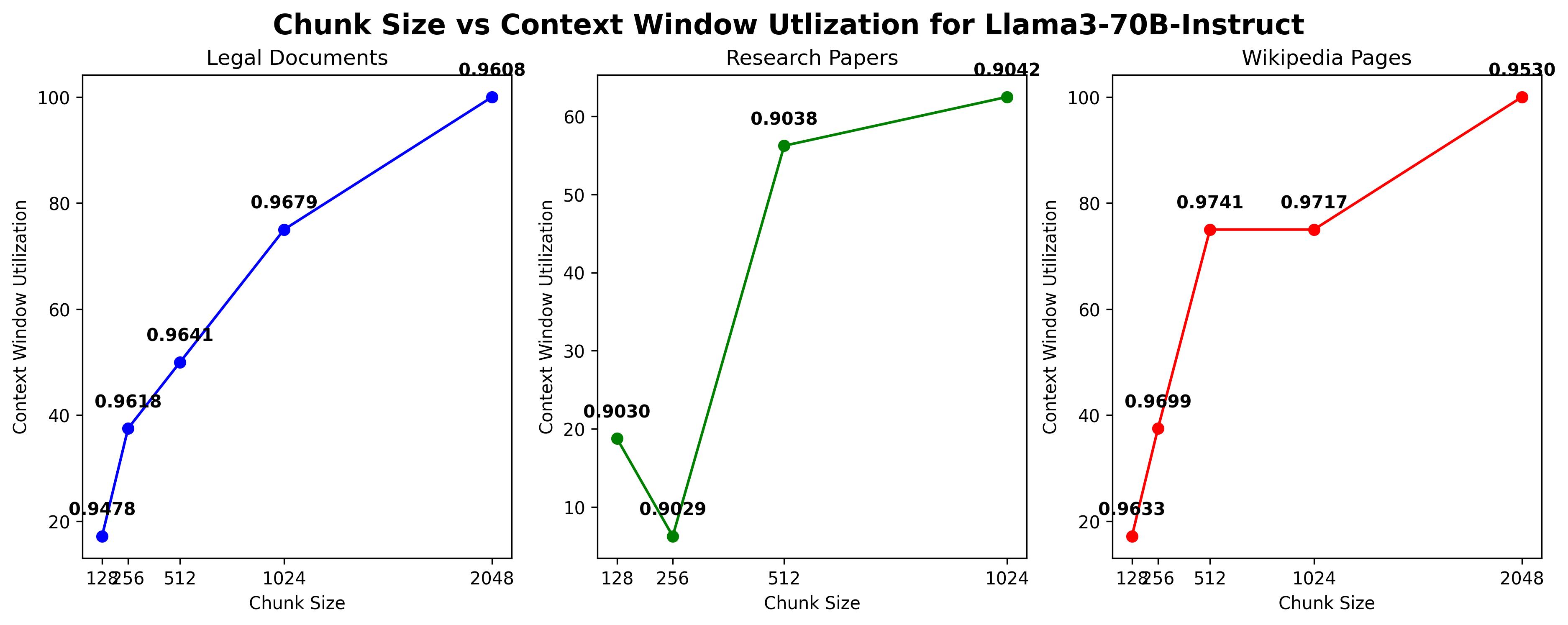}
  \caption{Chunk Size vs. Context Window Utilization across document types with annotated similarity scores for Llama3-70B-Instruct}
  \label{fig: Fig.3}
\end{figure*}

\begin{figure*}[h!]
  \centering
  \includegraphics[width=\textwidth]{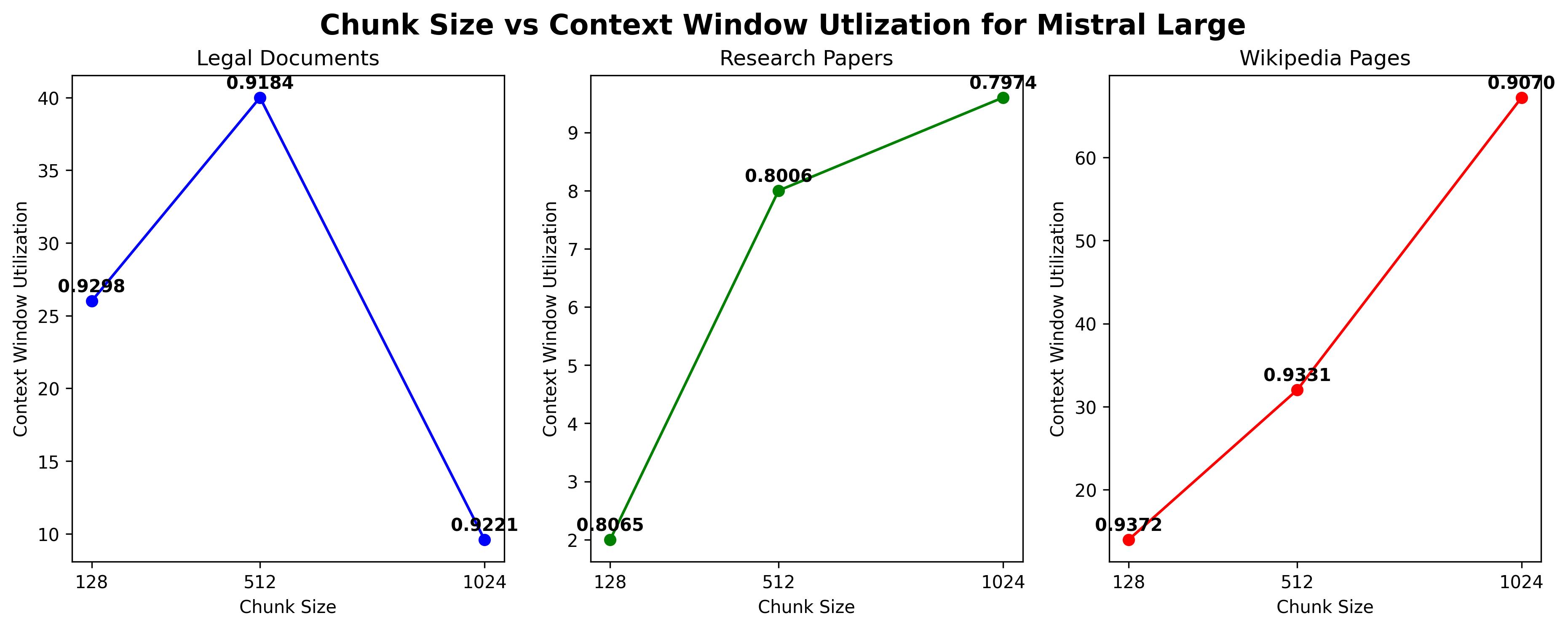}
  \caption{Chunk Size vs. Context Window Utilization across document types with annotated similarity scores for Mistral Large 2}
  \label{fig: Fig.4}
\end{figure*}
\cref{fig: Fig.3} illustrates the context window employed by Llama3-70B-Instruct to maximize similarity across various chunk sizes. Despite utilizing high chunk sizes such as 2048, the whole context window, there is no notable enhancement in the similarity scores. Reducing the size of the chunks leads to improved outcomes. Even when the chunk size is increased to 1024, the performance improvement compared to a chunk size of 512 is only somewhat higher. Additionally, there is a large increase in the use of the context window, or an identical level of context window utilization. This indicates that using a chunk size of 1024 improves performance when there are more chunks. However, accessing more chunks also increases the retrieval and response time, resulting in higher latency. A chunk size of 512 can provide developers with good results while reducing the need for retrieving several pieces, hence lowering latency. 

To assess the uniformity of the results among various LLMs, we performed an experiment with Mistral Large, which was released in February 2024 and has a context window of 32K tokens \cite{mistralLarge}. Given that chunk sizes of 512 and 1024 have shown better results, we evaluated chunk size of 128 in comparison to these two. In this instance too, it is evident that smaller context windows and suitable chunk sizes result in higher similarity scores. Furthermore, even when using Mistral Large, increasing the use of the context window does not enhance the similarity scores (\cref{fig: Fig.4}). This suggests that there exists an ideal proportion of LLM's context window which shall be utilized while performing RAG.

There have been other experiments which point the shortcomings of Long Context LLMs:
\begin{itemize}
    \item \textbf{LongICLBench: }The authors curate a dataset with increasing levels of difficulty and find that there is a dramatic degradation in performance of LLMs as task gets difficult and context length increase. Most state of the art LLMs are capped between easy to medium difficulty level.\cite{li2024longcontextllmsstrugglelong}
    \item \textbf{Same Task, More Tokens: }The authors isolate the effect of input length using multiple versions of the same sample, each being extended with padding of different lengths, types and locations. The findings show a notable degradation in LLMs' reasoning performance at much shorter input lengths than their technical maximum.\cite{levy2024tasktokensimpactinput}
    \item \textbf{BABILong: }It is a novel generative benchmark for evaluating the performance of NLP models in processing arbitrarily long documents with distributed facts. Their evaluations show that popular LLMs effectively utilize only 10-20\% of the context and their performance declines sharply with increased reasoning complexity. Among alternatives to in-context reasoning, Retrieval-Augmented Generation methods achieve a modest 60\% accuracy on single-fact question answering, independent of context length. \cite{kuratov2024babilongtestinglimitsllms}
    \item \textbf{Long Context RAG Performance of LLMs: }This recent blog post by Databricks Mosaic Research is extremely similar to our experimental setup. They find out saturation point various models (in terms of tokens) like 16k for gpt-4-turbo and claude-3-sonnet, 4k for mixtral-instruct and 8k for dbrx-instruct. The low saturation point of 4k token for mixtral-instruct could also be the reason behind inconsistency in our experiments as explained in section \ref{resultsanddiscussion}. They also classify what kind of errors LLMs make in RAG as context length increases. But this study doesn't experiment with multiple chunk size and show how granularity affects RAG with Long Context LLMs.\cite{databricksLongContext}
    \item \textbf{RULER: }It is a synthetic benchmark for evaluating long-context language models containing diverse task categories, retrieval, multi-hop tracing, aggregation and question answering. They show that on their benchmark long context LLMs display an effective context length of half their claimed context length or less which also aligns with our results.\cite{hsieh2024rulerwhatsrealcontext}
\end{itemize}

The aforementioned investigations highlight the limitations of long context language models (LLMs). Long context language models struggle to successfully make use of their claimed context length across different tasks and benchmarks. This presents a fresh avenue for us to investigate in order to determine the underlying reason, whether it is related to the model's architecture, the training process, or any other relevant factor. We have not yet discovered a method to ascertain the effective context length other than by testing it on various tasks and benchmarks.
\section{Significance and Conclusion}
Our study is an contribution towards understanding how effective state of the art LLMs are at utilizing their large context windows. As demonstrated by our findings, the right choice of chunk size  influences the length of context required to come up with best answers in RAG. Besides, more context does not help always. This  has important implications for LLM powered applications like VidyaRang \cite{harbola2024vidyarangconversationallearningbased} and CultureVo \cite{agarwala2024culturevogameutilizinggen}. The results in \cref{resultsanddiscussion} suggest a new hyper-parameter to be optimized for RAG which suggest  focusing on giving the LLM \textit{n} number of chunks that use a sufficient fraction of the context window rather than looking for \textit{top-k} similar chunks. We can put forward our recommendations based on the experiments thus far as follows:

\begin{itemize}
  \item \textbf{Optimization of Chunk Size:} Our experiments consistently show that chunk sizes of 512 and 1024 tokens deliver superior performance in terms of response quality across diverse datasets. 
  \item \textbf{Introduction of Context Window Utilization:}  Context Window Utilization is crucial as it ensures the efficient use of the context window.
  \item \textbf{Implications for Future Research:} Researchers have demonstrated that long context language models perform poorly when compared to RAG on different tasks\cite{phan2024ragvslongcontext,laban2024summaryhaystackchallengelongcontext}, it is yet to be determined what proportion of context window is genuinely useful and need to be explored further. Our work lays a foundation for developing more efficient RAG systems by focusing simultaneously on both chunk size and context window utilization.
\end{itemize}

\section{Limitations and Future Work}

We have only explored with three language models owing to time and resource restrictions. In the future, we want to examine using million token context window models to see how many tokens the LLM can effectively handle. The optimal utilization percentage for Llama3-70B-Instruct is 60-70\%. However, it is unclear if this proportion is consistent across all LLMs or varies depending on design. We would be investigating optimal value of context window utilization and its interplay with chunk size/top-k values for other open source LLMs in near future. We also wish to investigate if the under-utilization on the entire context length is caused my model architecture, training method or any such factors.

\section{Acknowledgement}
The authors thank I-Venture at Indian School of Business for infrastructural support toward this work.
Authors are extremely grateful to Prof. Bhagwan Chowdhry, Faculty Director (I-Venture at ISB) for his
continued encouragement, ideas and support to carry out this research.

\bibliographystyle{ACM-Reference-Format}
\bibliography{sources}

\appendix
\section*{Appendix}
\section{Custom Dataset}
\label{A}
On the following page we have presented few examples of question answer pairs from our dataset created using academic papers, legal documents, and Wikipedia articles. This question answer dataset is created by instructing GPT-4 Turbo.
\onecolumn
\begin{longtable}{@{\extracolsep{\fill}}|l|p{6.5cm}|p{6.5cm}|@{}}
\caption{Following table presents few examples of question answer pairs selected from our dataset  } \label{tab:appendix_dataset}\\
\toprule
Dataset & Question & Answer \\ \midrule
\endfirsthead

\multicolumn{3}{c}%
{\textit{Continued from previous page}} \\
\toprule
Dataset & Question & Answer \\ \midrule
\endhead

\midrule
\multicolumn{3}{r}{\textit{Continued on next page}} \\
\endfoot

\bottomrule
\endlastfoot

\multirow{2}{*}{Ours - Wikipedia} & How does the archaeological evidence suggest that early farming communities in Afghanistan compare to other regions in terms of historical significance?
 & The archaeological evidence suggests that early farming communities in Afghanistan are among the earliest in the world, dating back at least 7,000 years. Excavations of prehistoric sites indicate that humans were living in what is now Afghanistan at least 50,000 years ago. Artifacts from various historical periods, including the Paleolithic, Mesolithic, Neolithic, Bronze, and Iron Ages, have been found in Afghanistan.

In terms of historical significance, Afghanistan's early farming communities are comparable to regions like Egypt. The region is considered to be an important site of early historical activities, with archaeological sites that hold significant historical value. The urban civilization in Afghanistan is believed to have begun as early as 3000 BCE, and the country has been part of major ancient civilizations like the Indus Valley Civilization.

The archaeological evidence in Afghanistan, including the findings of artifacts and early settlements, indicates a long history of human habitation and cultural development. This historical significance places Afghanistan among the regions with early and important contributions to human civilization and cultural heritage. \\ 
& How did the Kushan Empire contribute to the spread of Buddhism in Afghanistan during the first century CE?
 & "The Kushan Empire, centered in Afghanistan, played a significant role in the spread and patronage of Buddhism in the region during the first century CE. Here are some ways in which the Kushan Empire contributed to the spread of Buddhism in Afghanistan:

1. **Patronage of Buddhist Culture**: The Kushan rulers were great patrons of Buddhist culture, supporting the development of Buddhist monasteries, stupas, and art. They provided financial and political support to Buddhist institutions, which helped Buddhism flourish in the region.

2. **Artistic and Architectural Contributions**: The Kushans were known for their artistic and architectural achievements, which included the creation of Buddhist sculptures, Gandharan art, and Buddhist monastic complexes. These artistic expressions helped in the visual representation and propagation of Buddhist teachings.

3. **Religious Tolerance**: The Kushan Empire was characterized by religious tolerance, allowing Buddhism to coexist and thrive alongside other belief systems in the region. This environment of religious pluralism facilitated the spread of Buddhism among diverse populations.

4. **Trade and Cultural Exchange**: The Kushan Empire was strategically located along the Silk Road, a major trade route connecting East and West. This facilitated cultural exchange and the transmission of Buddhist teachings to different regions, including Afghanistan.

5. **Syncretism and Adaptation**: The Kushans promoted syncretism between Buddhist beliefs and local traditions, making Buddhism more accessible and adaptable to the diverse cultural landscape of Afghanistan. This blending of religious practices helped Buddhism resonate with a wider audience.

Overall, the Kushan Empire's patronage of Buddhist culture, artistic contributions, religious tolerance, trade networks, and syncretic practices all contributed to the spread and flourishing of Buddhism in Afghanistan during the first century CE."
 \\ 
\multirow{2}{*}{Ours - Legal Document} & Why does the appellate judge find Mr. Vaidyanathan's submission regarding the comparison of wood vinegars in skin care preparations insufficient in the case of Diamond Star Global Sdn Bhd vs Joint Controller?
 & "The appellate judge finds Mr. Vaidyanathan's submission regarding the comparison of wood vinegars in skin care preparations insufficient in the case of Diamond Star Global Sdn Bhd vs Joint Controller for the following reasons:

1. **Lack of Comparative Analysis**: Mr. Vaidyanathan argued that the appellant's application was deficient because it did not compare the wood vinegar obtained from the Rhizophora apiculata tree with other wood vinegars used in skin care preparations to demonstrate superiority. However, the judge notes that the appellant's invention focused on specific properties of the wood vinegar from the Rhizophora apiculata tree, such as being free of guaiacol and having antimicrobial properties, which were not addressed in existing wood vinegars.

2. **Unique Properties**: The judge highlights that the appellant's application emphasized the unique properties of the wood vinegar from the Rhizophora apiculata tree, such as its ability to selectively target harmful bacteria while retaining beneficial lactic acid bacteria. These specific properties were not necessarily comparable to other wood vinegars used in skin care preparations, making a direct comparison inadequate to assess the inventive step of the appellant's invention.

3. **Inventive Step**: The judge emphasizes that the appellant's application demonstrated inventive step by identifying the unique properties of the wood vinegar from a specific tree, determining the optimal concentration for skin care preparations, and showcasing the antimicrobial benefits. This inventive step was based on the specific characteristics of the wood vinegar from the Rhizophora apiculata tree, rather than a generic comparison with other wood vinegars in the market.

Overall, the judge concludes that Mr. Vaidyanathan's argument regarding the comparison of wood vinegars in skin care preparations was insufficient to undermine the inventive step and uniqueness of the appellant's invention, which focused on distinct properties and applications of the wood vinegar from the Rhizophora apiculata tree."
 \\ 
& Why does the court find it incorrect to allege that the plaintiff obtained Suit Patents by committing fraud on the Indian Patent Office?
 & "The court finds it incorrect to allege that the plaintiff obtained Suit Patents by committing fraud on the Indian Patent Office for several reasons, as indicated in the context provided:

1. **Compliance with Section 8**: The plaintiff has asserted that it complied with Section 8 of the Patents Act by providing all relevant details and information related to foreign prosecution history or foreign patents/filings concerning the suit patents. The plaintiff maintains that it did not withhold any material information during the patent application process or examination in India.

2. **Absence of Evidence**: The court notes that the defendant's allegation of fraud is based on the assertion that the plaintiff secured patents over unpatentable subject matter with unduly broad claims. However, the court finds that the defendant has not provided substantial evidence to support these claims of fraud or misconduct on the part of the plaintiff during the patent application process.

3. **Challenge to Validity**: The defendant's objections and revocation petitions challenging the validity of the Suit Patents do not automatically imply fraud on the part of the plaintiff. The court emphasizes that the validity of a patent must be established before addressing infringement issues, but the mere challenge to validity does not necessarily indicate fraudulent conduct by the patentee.

4. **Prior Knowledge**: The court highlights that the defendant had prior knowledge of the plaintiff's patents for several years before raising objections to their validity. The defendant's delay in questioning the patents and only challenging them when the plaintiff asserted their legal rights against third parties weakens the argument of fraud or misconduct by the plaintiff.

5. **Lack of Serious Challenge**: The court observes that the defendant's objections do not present a serious challenge to the validity of the Suit Patents. The defendant's conduct and actions, including delays in raising objections and lack of substantial evidence supporting fraud allegations, do not convincingly demonstrate fraudulent behavior on the part of the plaintiff.

In summary, the court finds it incorrect to allege that the plaintiff obtained Suit Patents by committing fraud on the Indian Patent Office due to the lack of substantial evidence, the plaintiff's compliance with legal requirements, the defendant's prior knowledge of the patents, and the absence of a serious challenge to the patents' validity."
 \\ 
\multirow{2}{*}{Ours - Research Papers} & How do PINNs allow for query at any temporal/spatial location within a given domain once trained?
 & "Physics-Informed Neural Networks (PINNs) allow for queries at any temporal/spatial location within a given domain once trained through the following mechanisms:

1. **Implicit Function Representation**:
   - PINNs are trained to learn the underlying physics of a system and represent the solution as an implicit function. This implicit function can be evaluated at any point within the domain, allowing for queries at arbitrary temporal and spatial locations.

2. **Generalization**:
   - PINNs generalize the learned physics across the entire domain. Once trained on a dataset, the network can provide accurate predictions at new, unseen locations within the domain without the need for additional training data at those specific points.

3. **Interpolation**:
   - PINNs use interpolation techniques to estimate the solution at points where data is not available. By learning the underlying physics and relationships in the data, PINNs can interpolate between known data points to provide predictions at arbitrary locations.

4. **Continuous Function Approximation**:
   - PINNs approximate the solution as a continuous function, allowing for smooth predictions across the domain. This continuous representation enables querying at any point, ensuring consistent and accurate results throughout the spatial and temporal domain.

5. **Efficient Inference**:
   - Once trained, PINNs offer efficient inference capabilities, making it computationally feasible to query the model at multiple temporal and spatial locations. This efficiency allows for quick and accurate predictions at any point of interest within the domain.

By leveraging these characteristics, PINNs enable flexible"
 \\ 
& What is the significance of the two-network architecture proposed for solving problems with moving interfaces, as outlined in the text?
 & "The two-network architecture proposed for solving problems with moving interfaces, as outlined in the text, holds significant importance for several reasons:

1. **Complex Problem Solving**:
   - The two-network architecture allows for solving complex problems involving moving interfaces, such as fluid-structure interaction or free surface flows, by dividing the task into two sub-problems that can be addressed independently.

2. **Interface Tracking**:
   - One network in the architecture focuses on tracking the moving interface, which is crucial for problems where the interface location changes with time. This dedicated network ensures accurate representation and tracking of the interface dynamics.

3. **Solution Reconstruction**:
   - The second network is responsible for learning the solution of the underlying partial differential equations (PDEs) with appropriate constraints at the interface. This network reconstructs the complete solution field based on the interface information provided by the first network.

4. **Improved Generalization**:
   - By separating the interface tracking and solution reconstruction tasks, the architecture can achieve improved generalization and accuracy in solving problems involving dynamic interfaces."
 \\ 

\end{longtable}

\end{document}